\documentclass[11pt]{article}

\usepackage[utf8]{inputenc}
\usepackage[T1]{fontenc}
\usepackage[english]{babel}
\usepackage{lmodern}
\usepackage[margin=0.9in]{geometry}
\usepackage{graphicx}
\usepackage{booktabs}
\usepackage{tabularx}
\usepackage{array}
\usepackage{amsmath}
\usepackage{amssymb}
\usepackage{xurl}
\usepackage[hidelinks,hypertexnames=false]{hyperref}
\usepackage{enumitem}
\usepackage[section]{placeins}
\usepackage{float}
\usepackage{microtype}
\usepackage{xcolor}
\usepackage{caption}
\setlist[itemize]{leftmargin=*,topsep=3pt,itemsep=2pt}
\setlist[enumerate]{leftmargin=*,topsep=3pt,itemsep=2pt}
\definecolor{traceframe}{RGB}{84,103,128}
\definecolor{traceback}{RGB}{246,248,251}
\definecolor{tracelabel}{RGB}{49,72,103}
\newcommand{\schwartz}{Schwartz-10}
\newcommand{\val}[1]{\textsc{#1}}
\newcommand{\acc}{Acc@1}
\newcommand{\accthree}{Acc@3}

\hypersetup{
  pdftitle={Which Values Do LLMs Confuse? A Schwartz-Based Recognition Study},
  pdfauthor={Andrei Chetvergov, Stepan Ukolov, Timofei Sivoraksha, Alexander Evseev, Mikhail Solovev, Valeriia Kuschenko, Maria Chistyakova, Sergey Bolovtsov}
}

\begin{document}
\raggedbottom

\begin{center}
{\LARGE\bfseries Which Values Do LLMs Confuse?\par}
\vspace{0.18em}
{\Large A Schwartz-Based Recognition Study\par}
\vspace{0.9em}

{\normalsize
Andrei Chetvergov\textsuperscript{1,2},
Stepan Ukolov\textsuperscript{1,2},
Timofei Sivoraksha\textsuperscript{1,2},
Alexander Evseev\textsuperscript{1,2}\\[0.25em]
Mikhail Solovev\textsuperscript{1,2},
Valeriia Kuschenko\textsuperscript{2},
Maria Chistyakova\textsuperscript{2},
Sergey Bolovtsov\textsuperscript{1,2}\par}
\vspace{0.75em}

{\small
\textsuperscript{1}Ivannikov Institute for System Programming of the Russian Academy of Sciences, Moscow, Russia\\[0.12em]
\textsuperscript{2}Russian Presidential Academy of National Economy and Public Administration, Moscow, Russia\par}
\vspace{0.55em}

{\footnotesize
\texttt{\{chetvergov-as,ukolov-sd,sivoraksha-ta\}@ranepa.ru}\\[-0.05em]
\texttt{\{aevseev-23-01,mchistyakova-25,bolovtsov-sv\}@ranepa.ru}\\[-0.05em]
\texttt{\{msolovev-24,vkuschenko-22\}@edu.ranepa.ru}\par}
\end{center}
\vspace{0.35em}

\begin{abstract}
Large language models are increasingly evaluated through the values they endorse, but such evaluations presuppose that models can identify the value expressed in a concrete situation. We study this prerequisite as controlled top-1 recognition over Schwartz's ten basic values. Our evaluation set contains 1,000 Russian situational texts, balanced across the ten values and independently labeled by two human annotators per item. We evaluate 21 instruction-tuned LLM runs under a fixed ranked-response protocol; 20 runs with reliable outputs form the semantic panel. Pooled \acc{} is 0.683 and \accthree{} is 0.892, showing that models often locate the correct motivational region while ranking close alternatives unstably. Adjacent values account for 50.9\% of semantic errors, compared with 24.4\% under a checkpoint-specific null. Eight directed confusions recur across checkpoints and human-confirmed subsets. Several are strongly asymmetric---including \val{Universalism}$\rightarrow$\val{Benevolence}, \val{Tradition}$\rightarrow$\val{Conformity}, and \val{Security}$\rightarrow$\val{Power}---whereas \val{Stimulation}--\val{Hedonism} forms a bidirectional boundary. Their severity is checkpoint-specific and can bias higher-order value profiles. The results motivate value-recognition evaluation that combines exact accuracy, ranked recovery, and directed error analysis.
\end{abstract}

\section{Introduction}

Human values are increasingly used to evaluate large language models (LLMs): systems are compared by value orientations, moral choices, pluralistic judgments, and value-informed behavior. These evaluations depend on a simpler capability that is rarely isolated. Before asking whether a model endorses or follows a value, one should establish whether it can recognize which value is expressed in ordinary language.

Recognition errors matter because predicted value labels can become features for profiling, routing, recommendation, policy analysis, or personalized intervention. A systematic \val{Universalism}$\rightarrow$\val{Benevolence} shift can reduce a demand for equal access to a story of charitable care; \val{Conformity}$\rightarrow$\val{Security} can turn normative compliance into threat avoidance; and \val{Security}$\rightarrow$\val{Power} can read protective action as domination. These examples make the direction of an error more informative than a generic accuracy loss.

\paragraph{Schwartz-10 in brief.}
Schwartz's circumplex arranges ten values so that neighboring positions express compatible motives and opposite positions express competing priorities~\cite{schwartz1992universals,schwartz2012refining}. Openness to Change comprises \val{Self-Direction} (independent thought and action), \val{Stimulation} (novelty and excitement), and \val{Hedonism} (pleasure); Self-Enhancement comprises \val{Achievement} (success through competence) and \val{Power} (status and control); Conservation comprises \val{Security} (safety and stability), \val{Conformity} (restraint under social norms), and \val{Tradition} (respect for inherited customs); Self-Transcendence comprises \val{Benevolence} (welfare of close others) and \val{Universalism} (welfare of all people and nature). Recognition therefore requires both locating the broad motivational region and distinguishing nearby motives. A single accuracy score can conceal the more diagnostic question posed by our title: \emph{which values does a model confuse, in which direction, and more often than expected from its general label preferences?}

\begin{figure}[H]
\centering
\includegraphics[width=0.96\textwidth]{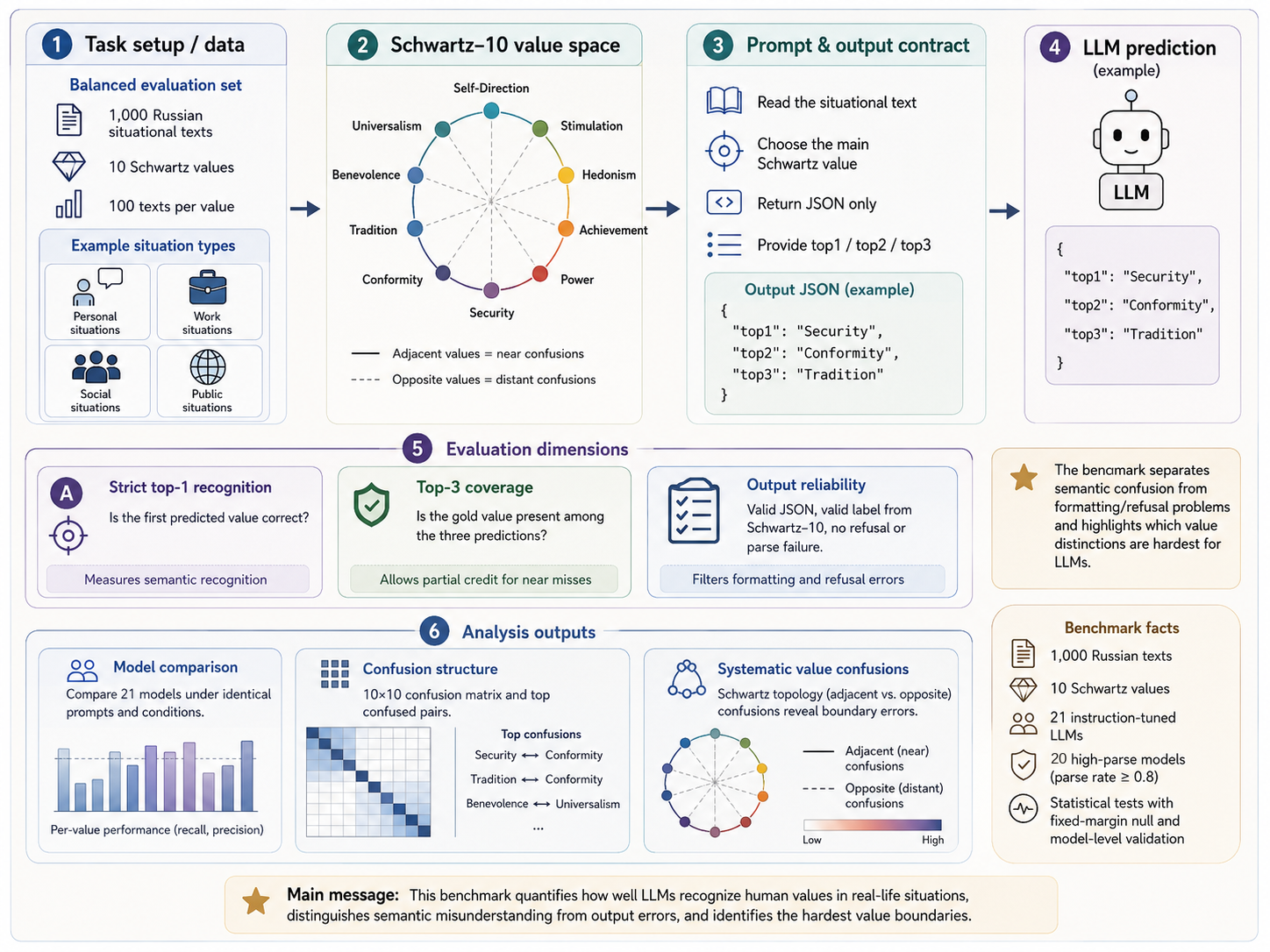}
\caption{Task overview. A model maps a situational text to one primary Schwartz value and two alternatives under a fixed ranked-response schema. Evaluation separates output reliability, top-1 recognition, top-3 coverage, and confusion structure.}
\label{fig:overview}
\end{figure}

We study primary-value recognition in short Russian situational texts. Each model returns a ranked prediction over \schwartz{}. Russian provides the empirical setting; the object of analysis is the structure of value recognition and confusion. Figure~\ref{fig:overview} summarizes the task.

The study addresses three research questions:
\begin{itemize}
    \item \textbf{RQ1 -- Recognition and ranked recovery.} How accurately do instruction-tuned LLMs recognize primary Schwartz values, which values are hardest, and how often does the reference remain at ranks 2--3 after a top-1 disagreement?
    \item \textbf{RQ2 -- Directed confusion and checkpoint heterogeneity.} Which directed boundaries exceed checkpoint-specific label-preference baselines, reproduce across checkpoints and family exclusions, and form distinct model-specific error fingerprints?
    \item \textbf{RQ3 -- Human robustness and profile distortion.} Do the conclusions persist on independently confirmed items, what semantic mechanisms recur, and how do fine-grained errors alter higher-order motivational profiles?
\end{itemize}

Our contributions are fourfold. First, we formulate primary-value recognition as a controlled diagnostic task and release a balanced set of 1,000 unique Russian texts with two independent human labels for every item. Second, we separate output reliability, exact recognition, and ranked recovery, and introduce checkpoint-specific fixed-margin inference with replication and leave-one-family-out tests. Third, we characterize eight reproducible directed transitions, expose checkpoint-specific confusion fingerprints, and audit all 62 high-consensus cases for recurring semantic mechanisms. Fourth, we repeat the central analyses on two human-confirmed subsets and quantify the resulting distortion of higher-order value profiles over 20,000 model--text observations.

\section{Related Work}

\paragraph{Value Recognition in Text.}
Webis-ArgValues-22 established value inference from natural-language arguments as an NLP task~\cite{kiesel2022identifying}; ValueEval and the Touche23-ValueEval extension enabled system and per-category comparison over larger multilabel argument collections~\cite{kiesel2023semeval,mirzakhmedova2024touche}. These resources establish values as operational textual categories. Our study adds a compact psychological geometry in which directed boundaries and circumplex distance are directly interpretable.

\paragraph{LLM Value and Scenario Benchmarks.}
ValueBench evaluates orientations and value understanding across many psychometric dimensions~\cite{ren2024valuebench}; Value Kaleidoscope / ValuePrism emphasizes pluralistic rights and duties in situations~\cite{sorensen2024value}; Generative Psychometrics and ValueLlama support scalable value measurement from free text~\cite{ye2025measuring}. INVP studies value priorities in social decisions~\cite{liu2025invp}, ValueActionLens examines the value--action gap~\cite{shen2025valueaction}, and Value Portrait links realistic interactions to human value scores~\cite{han2025portrait}. We isolate a prerequisite shared by these settings: distinguishing the expressed value before interpreting a profile or action.

\paragraph{Annotation Uncertainty and Russian Sources.}
Value expression in natural text is difficult to annotate consistently. Epstein et al. report substantial human and model variation in social-media value ratings~\cite{epstein2025measuring}; Milkova and Rudnev emphasize calibrated non-English Schwartz annotation, expert verification, and error-structure analysis~\cite{milkova2026measuring}. We therefore combine exact agreement between two construction models with complete two-annotator human validation. The candidate pool draws on Russian sensitive-topic and appropriateness data~\cite{babakov2021inappropriate} and TAPE ethics examples~\cite{taktasheva2022tape}; source labels select candidates only. Scenario resources such as ETHICS, Social Chemistry 101, and Moral Stories motivate the use of concrete social situations~\cite{hendrycks2021ethics,forbes2020social,emelin2021moral}.

\section{Task and Dataset}

\subsection{Recognition Task and Response Schema}

For a text $x$, the model returns a ranking $(\hat y_1,\hat y_2,\hat y_3)$ over the ten Schwartz values, and the primary task is $x\mapsto\hat y_1$. The judgment concerns the value expressed in the situation---not the model's own preference and not whether the situation is morally acceptable. Label names are presented without definitions or demonstrations. Empty responses, refusals, and responses without an admissible top-1 label are output failures; semantic metrics use the extracted ranking, while rationales support the qualitative audit.

\begin{center}
\includegraphics[width=0.96\textwidth]{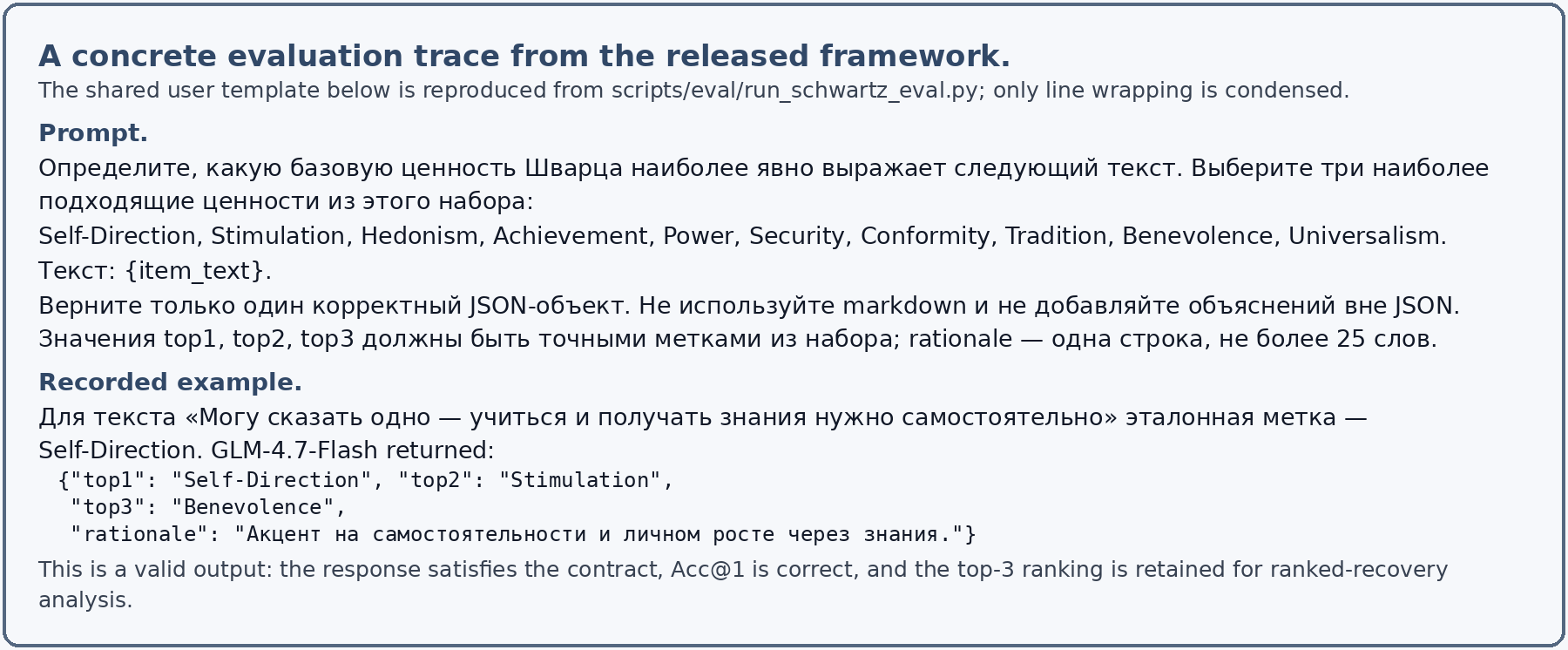}
\end{center}

\subsection{Candidate Pool and Construction Labels}

The candidate pool contains 70,649 Russian texts from four source slices: 33,303 sensitive-topic texts, 28,931 high-confidence inappropriate messages, a 5,000-item high-confidence appropriate sample, and 3,415 TAPE ethics examples. Source categories are used solely for candidate selection; all final labels belong to the Schwartz-10 space.

\begin{figure}[H]
\centering
\includegraphics[width=0.98\textwidth]{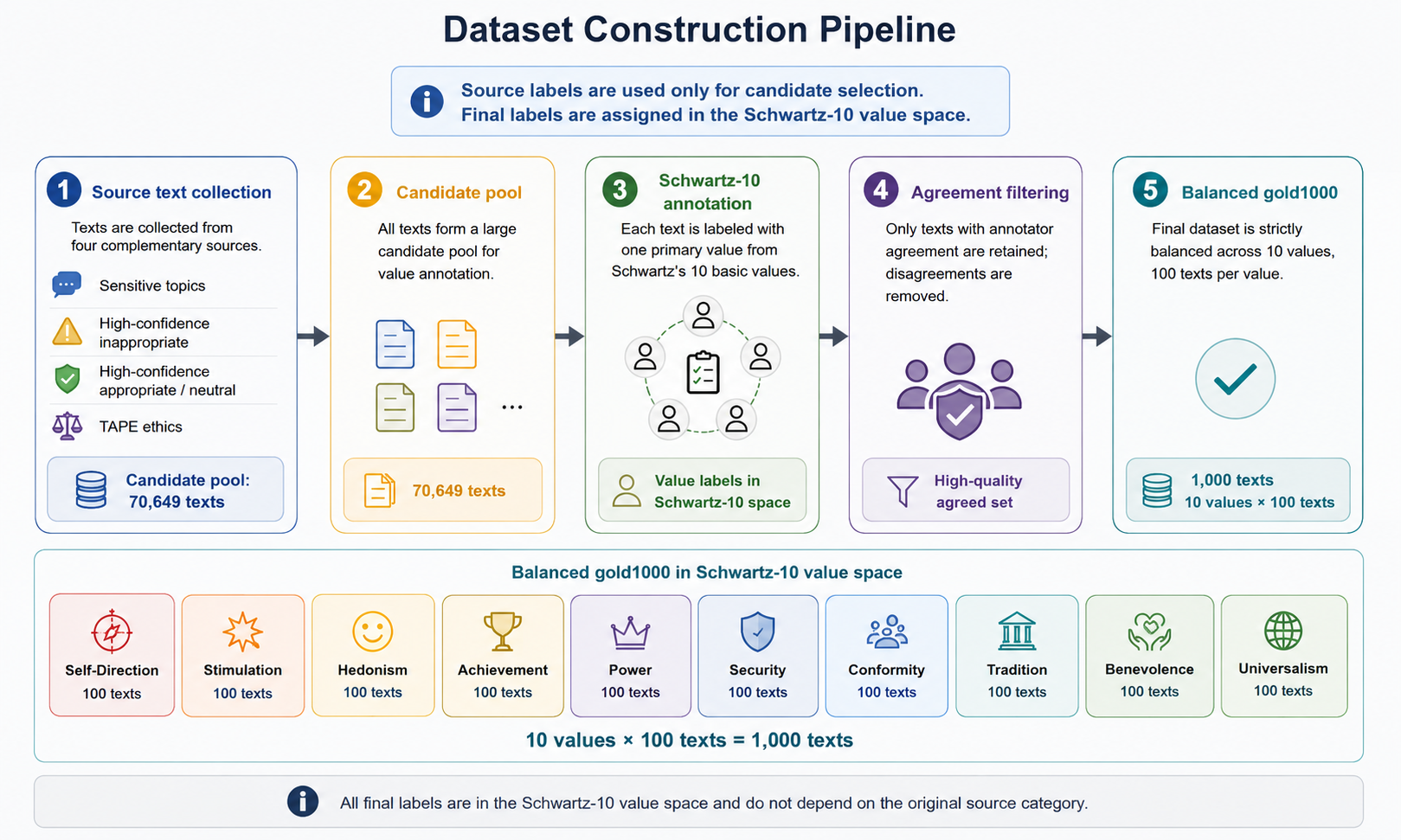}
\caption{Dataset construction. Original source categories select candidates only. Reference labels are assigned in the Schwartz-10 space by two separate construction models and retained under exact top-1 agreement. Every selected item is then independently labeled by two humans.}
\label{fig:pipeline}
\end{figure}

The construction stage is shown in Figure~\ref{fig:pipeline}. A specialized ValueLlama annotator labels the candidates; 43,165 items are cross-annotated by GPT-4.1-mini with an explicit \texttt{none} option. Exact top-1 agreement yields 14,516 eligible candidates. For each value, the 100 highest-scoring agreed items are selected subject to a maximum of 60 items from any one source slice. The final set contains 1,000 unique text surfaces, exactly 100 per value: 521 sensitive-topic, 306 high-confidence inappropriate, 85 high-confidence appropriate, and 88 TAPE ethics texts.

\begin{table}[H]
\caption{Construction-stage selectivity by ValueLlama label. ``Agree'' is exact top-1 agreement with GPT-4.1-mini; \texttt{none} is the GPT abstention rate. The final set then selects 100 items per value.}
\label{tab:constructionaudit}
\centering
\scriptsize
\setlength{\tabcolsep}{4.0pt}
\renewcommand{\arraystretch}{0.94}
\begin{tabular}{@{}lrrrr@{}}
\toprule
Value & Cross-annotated & Agree rate & GPT \texttt{none} & Eligible \\
\midrule
Self-Direction & 3,045 & .200 & .173 & 609 \\
Stimulation & 4,463 & .192 & .312 & 855 \\
Hedonism & 8,487 & .324 & .237 & 2,749 \\
Achievement & 3,748 & .356 & .177 & 1,333 \\
Power & 1,624 & .576 & .132 & 936 \\
Security & 10,288 & .402 & .124 & 4,136 \\
Conformity & 3,622 & .271 & .246 & 981 \\
Tradition & 3,168 & .551 & .131 & 1,746 \\
Benevolence & 1,494 & .284 & .177 & 425 \\
Universalism & 3,226 & .231 & .155 & 746 \\
\bottomrule
\end{tabular}
\end{table}

\subsection{Independent Human Validation}

\begin{figure}[H]
\centering
\fbox{\includegraphics[width=0.76\textwidth]{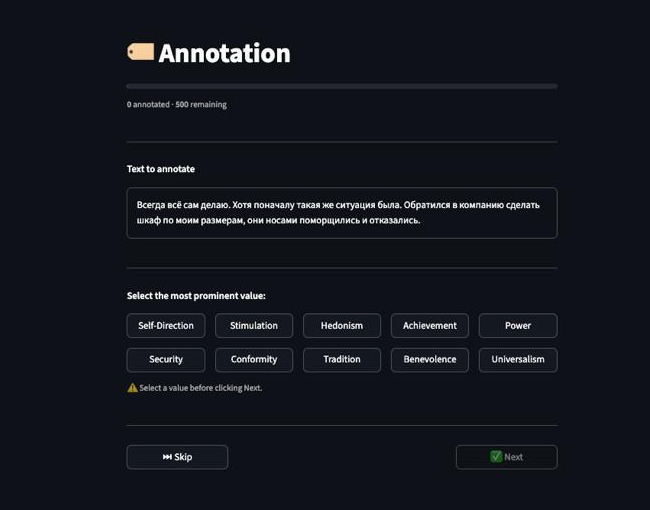}}
\caption{Human annotation interface. Annotators viewed one Russian situational text at a time and selected its most prominent value from the ten Schwartz labels.}
\label{fig:annotationui}
\end{figure}

Twelve annotators worked independently in a dedicated web interface (Figure~\ref{fig:annotationui}). Before annotation, they received a concise introduction to Schwartz's basic value theory and a reference guide with definitions and illustrative examples for each of the ten values; the guide remained available during annotation. Each screen presented one Russian text and the ten Schwartz labels as a forced primary-value choice. Two labels were collected for every item, yielding 2,000 judgments with complete coverage.

The label jointly selected by the two construction models receives independent support from at least one human for 950 of 1,000 items (95.0\%, Wilson 95\% CI [93.5, 96.2]) and from both humans for 611 items (61.1\%). Across individual judgments, human-to-reference agreement is 78.1\% with Cohen's $\kappa=.756$; at-least-one-human support ranges from 88\% to 99\% across the ten values. Under the stricter criterion that both humans independently choose exactly the same forced primary label, pair agreement is 62.4\% ($\kappa=.582$). These measures provide broad independent support and preserve information about close boundaries that admit plausible secondary readings. We use \emph{reference label} for the primary label selected under this protocol.

\section{Experimental Protocol}

\subsection{Model Panel and Recognition Metrics}

We evaluate 21 instruction-tuned runs under the same prompt, item order, temperature setting, and parser, yielding 21,000 outputs. All runs enter output-reliability analysis. Semantic analyses use the 20 runs with output-validity rate at least 0.8; the excluded run has rate 0.577 and the next-lowest rate is 0.948, so thresholds of 0.8 and 0.9 select the same panel.

\paragraph{Metric guide.}
Four related measures answer different questions. The \emph{output-validity rate} (reported as Parse in figures) is the fraction of items with an admissible top-1 label. Strict end-to-end \acc{} counts every invalid output as incorrect:
\[
\mathrm{StrictAcc@1}=N^{-1}\sum_i \mathbf{1}[\mathrm{parse}_i\land \hat y_{i,1}=y_i].
\]
Strict \accthree{} asks whether the reference appears anywhere in the returned top three, again over all items. By contrast, \emph{top-3 rescue} is conditional: among valid top-1 errors only, it asks whether the reference remains at rank 2 or 3. We also report macro-F1, per-value precision and recall, and directed confusion counts.

\subsection{Directed-Confusion Inference}

Raw counts can reflect a checkpoint's general tendency to overuse particular labels. Our fixed-margin null is a label-bias-preserving baseline: within each checkpoint it reshuffles erroneous destinations while holding fixed both the number of errors from every reference value and the overall frequency of each wrong predicted label. We draw 10,000 such samples. For each of 90 directed transitions $a\rightarrow b$, a one-sided test asks whether the observed count exceeds this baseline; Holm correction is the multiple-testing adjustment used to control family-wise error.

\paragraph{Evidence guide.}
We use three increasingly strict levels. \emph{Aggregate significance} pools the evidence across all predictions. \emph{Checkpoint replication} asks whether observed-minus-null excess is consistently positive across checkpoints, using the Wilcoxon signed-rank test (a nonparametric paired test) with Holm correction. \emph{Leave-one-family-out robustness} repeats that replication test after removing each of 11 model-family groups. Finally, comparing $a\rightarrow b$ with $b\rightarrow a$ distinguishes two-way boundary ambiguity from one-way category collapse.

\subsection{Robustness, Controls, and Case Audit}

The full 1,000-item set is primary. We repeat model ranking and directed-transition inference on 950 items confirmed by at least one human and 611 confirmed by both. The ten values are also aggregated into four non-overlapping orientations: Openness to Change (Self-Direction, Stimulation, Hedonism), Self-Enhancement (Achievement, Power), Conservation (Security, Conformity, Tradition), and Self-Transcendence (Benevolence, Universalism). Assigning Hedonism to Openness is an explicit reporting convention.

Exact-correctness logit models over 20,000 checkpoint--text observations include checkpoint, reference-value, and source fixed effects plus standardized length; standard errors are clustered by text and checkpoint. A checkpoint-by-value interaction model---allowing each checkpoint to have its own difficulty profile across the ten values---tests whether one overall checkpoint-quality effect plus one overall value-difficulty effect is sufficient. For each core transition, a likelihood-ratio test compares models with and without checkpoint-specific transition rates after source and length controls. Finally, all 62 texts for which at least 10 of 20 checkpoints produce one of six core transitions are coded for salient cue, mechanism, ambiguity, and interpretation. Complete model identifiers, all 90 transition tests, row-level predictions, human labels, case codes, full regression coefficients, and reproduction scripts are available in the linked resources.

\section{Results}

\subsection{Recognition and Ranked Recovery}

\begin{figure}[H]
\centering
\includegraphics[width=0.98\textwidth]{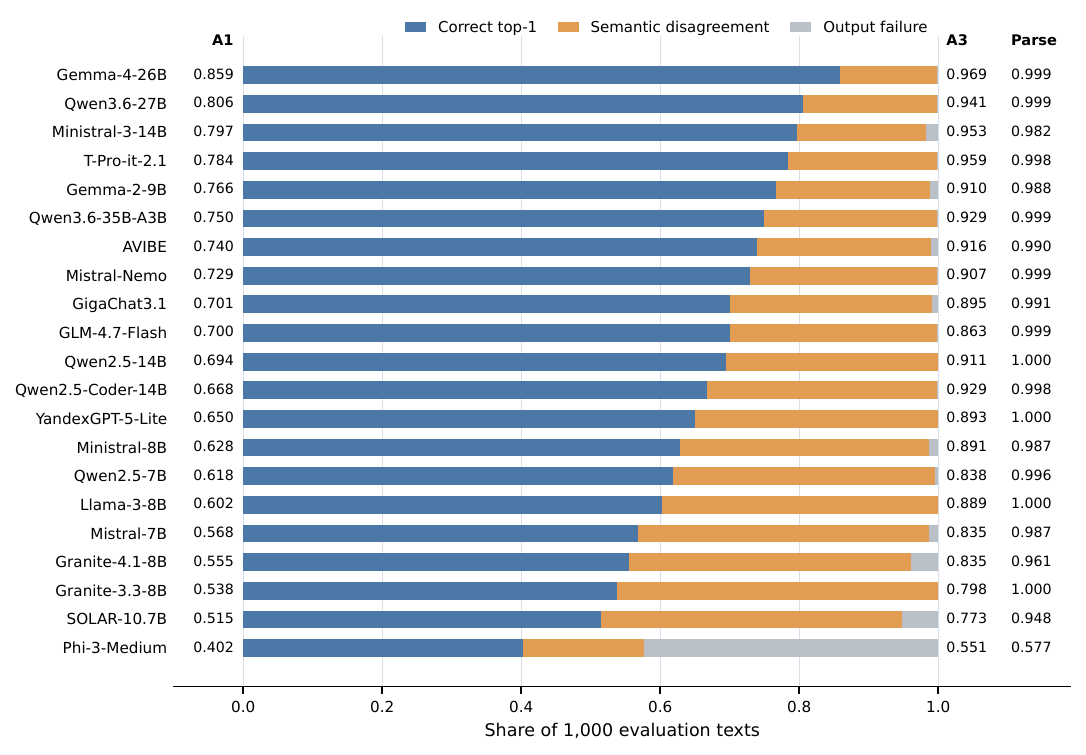}
\caption{End-to-end outcomes on 1,000 texts for all 21 runs. Each bar separates correct top-1 recognition, semantic disagreement, and output failure; numeric columns report strict Acc@1, strict Acc@3, and output-validity rate (Parse).}
\label{fig:modelperf}
\end{figure}

Figure~\ref{fig:modelperf} decomposes all 21 runs into correct top-1 recognition, semantic disagreement, and output failure. Gemma-4-26B reaches 0.859 strict \acc{} (95\% bootstrap CI [0.837, 0.880]), followed by Qwen3.6-27B at 0.806, Ministral-3-14B at 0.797, and T-Pro-it-2.1 at 0.784. Paired bootstrap differences separate Gemma from every other top-six system; intervals among the next three overlap.

\paragraph{Overall Performance and Ranked Recovery.}
Across the 20-model semantic panel, pooled strict \acc{} is 0.683 and \accthree{} is 0.892. Among 6,153 valid top-1 disagreements, 4,166 (67.7\%) recover the reference at rank 2 or 3. Rescue is much more common for adjacent than non-adjacent errors (87.5\% versus 47.2\%; bootstrap difference 40.4 percentage points, 95\% CI [37.3, 43.5]). This gap indicates that many local top-1 errors arise from unstable ordering among plausible neighbors while the broader motivational region remains correctly located. Twenty runs have output-validity rate at least 0.8; Phi-3-Medium is retained in the end-to-end comparison but excluded from semantic inference because its rate is 0.577.

\begin{figure}[H]
\centering
\includegraphics[width=0.98\textwidth]{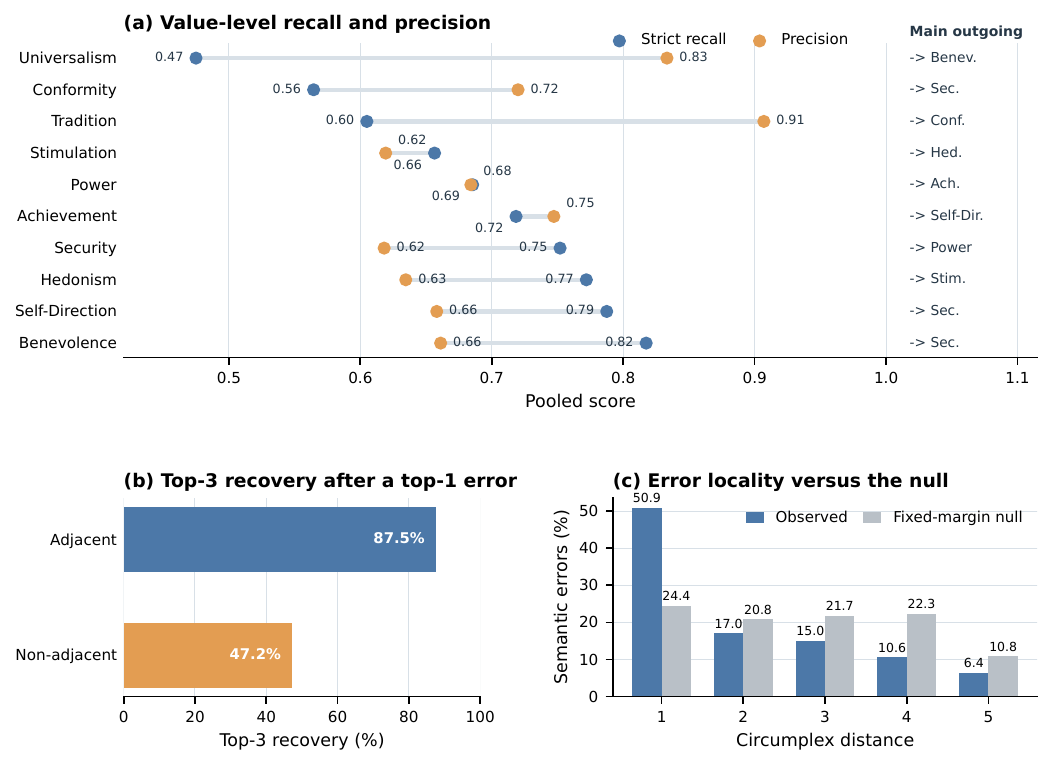}
\caption{Recognition diagnostics over the 20-model semantic panel. (a) Pooled strict recall and precision by value; arrows name the dominant erroneous destination. (b) Recovery of the reference value at ranks 2--3 after an adjacent or non-adjacent top-1 error. (c) Observed semantic-error share by circumplex distance versus the checkpoint-specific fixed-margin null.}
\label{fig:boundary}
\end{figure}

\paragraph{Value-Level Difficulty.}
Figure~\ref{fig:boundary}a makes the category-specific pattern visible without reducing it to one pooled score. Universalism has the lowest recall (0.475) but high precision (0.833), and Tradition shows the same selective pattern (recall 0.605, precision 0.907). They occupy only 5.8\% and 6.7\% of valid predictions despite 10\% reference prevalence, indicating selective under-use. Benevolence is recovered most often (recall 0.818), while the arrows identify the dominant outgoing boundary for every value. The controlled logit confirms this pattern: relative to Achievement, Universalism is less likely to be recovered (OR $=0.40$, 95\% CI [0.21, 0.77], $p=.006$), whereas Benevolence is more likely (OR $=2.18$, [1.16, 4.08], $p=.015$). Text length is unrelated to correctness ($p=.670$).

\FloatBarrier

\subsection{Directed Confusion and Checkpoint Heterogeneity}

\begin{figure}[H]
\centering
\includegraphics[width=0.84\textwidth]{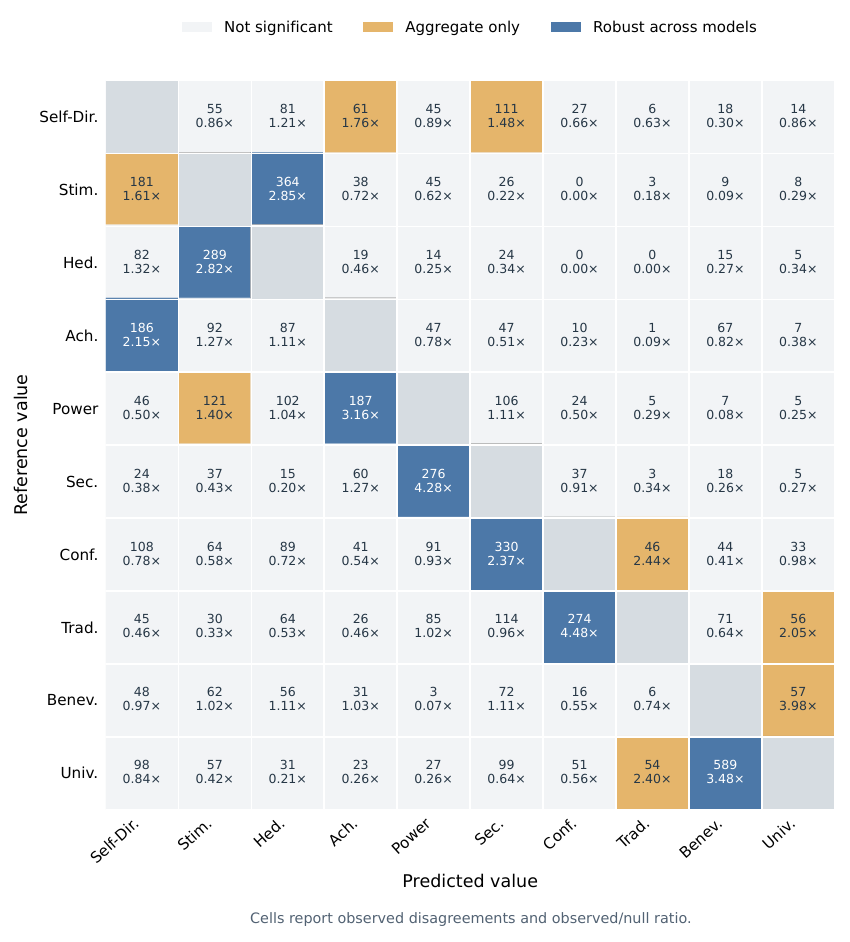}
\caption{Directed confusion inference. Each off-diagonal cell reports the observed count and observed-to-null ratio. Blue cells survive aggregate fixed-margin and checkpoint-level Holm correction; orange cells survive aggregate correction only.}
\label{fig:significance}
\end{figure}

\paragraph{Evidence levels and robust transitions.}
The fixed-margin test rejects random reassignment of erroneous labels ($p_{\mathrm{MC}}=0.0001$). At the pooled level, 16 transitions survive Holm correction; eight also replicate across checkpoints. The strongest enrichments are \val{Tradition}$\rightarrow$\val{Conformity} (4.48 times the null) and \val{Security}$\rightarrow$\val{Power} (4.28 times); the other six replicated transitions range from 2.15 to 3.48 times the null. Six remain significant after every model-family exclusion, and the remaining two after 10 of 11 exclusions. Figure~\ref{fig:significance} reports observed counts and all aggregate-significant cells.

\paragraph{Locality and direction.}
Locality and direction are separate findings. Adjacent values account for 50.9\% of 6,153 semantic disagreements versus 24.4\% under the null, while opposite values account for 6.4\% versus 10.8\% expected. \val{Stimulation}--\val{Hedonism} is a genuinely bidirectional boundary (18.2\% versus 14.5\% of the respective source values). By contrast, \val{Universalism}$\rightarrow$\val{Benevolence} (29.5\% versus 2.9\% in reverse), \val{Conformity}$\rightarrow$\val{Security} (16.5\% versus 1.9\%), and \val{Tradition}$\rightarrow$\val{Conformity} (13.7\% versus 2.3\%) indicate one-way category collapse.

\begin{figure}[H]
\centering
\includegraphics[width=0.98\textwidth]{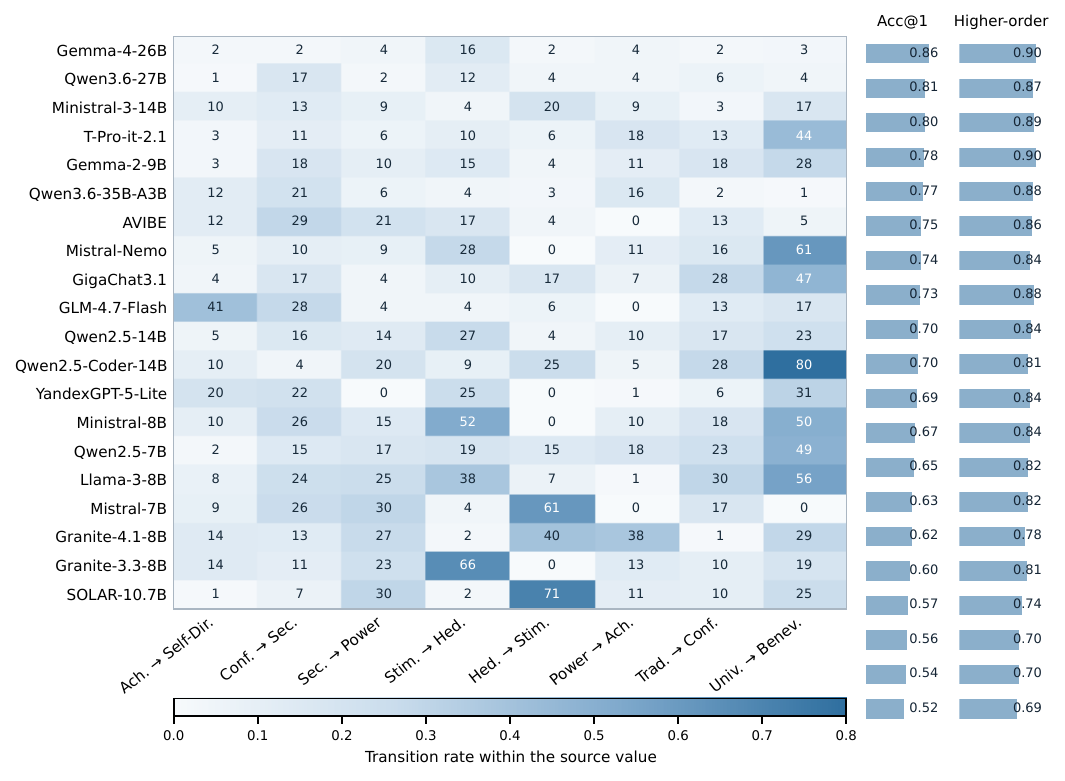}
\caption{Checkpoint-specific confusion fingerprints for the eight robust transitions. Cells are transition rates within the reference value (percent labels); side bars show strict Acc@1 and higher-order accuracy.}
\label{fig:modelheat}
\end{figure}

\paragraph{Checkpoint fingerprints.}
Figure~\ref{fig:modelheat} shows that shared boundaries do not imply uniform model behavior. Qwen2.5-Coder-14B maps \val{Universalism} to \val{Benevolence} on 80\% of Universalism items; SOLAR-10.7B favors \val{Hedonism}$\rightarrow$\val{Stimulation} on 71\%, whereas Granite-3.3-8B shows the reverse \val{Stimulation}$\rightarrow$\val{Hedonism} shift on 66\%. GLM-4.7-Flash maps \val{Achievement} to \val{Self-Direction} on 41\%. Gemma-4-26B keeps most of the same transitions comparatively small while leading overall accuracy.

The checkpoint-by-value interaction test asks whether one general model-quality effect and one general value-difficulty effect explain the whole matrix. They do not ($\chi^2(171)=3499.0$, $p<.001$), and all eight transition-specific heterogeneity tests survive Holm correction. Within-family and between-family profile similarities are statistically indistinguishable (mean cosine similarity 0.634 versus 0.621; permutation $p=.368$). The shared result concerns difficult boundaries; their severity and direction remain checkpoint-specific.

\subsection{Human Robustness, Interpretation, and Profile Distortion}

The independently confirmed subsets produce slightly stronger recognition results. Table~\ref{tab:robustprofile} shows higher Acc@1 and Acc@3, virtually unchanged checkpoint ranking, and retention of all eight full-set transitions on both subsets. The transition structure therefore persists under independent human confirmation.

\begin{table}[H]
\caption{Human-confirmed sensitivity and higher-order profile distortion. Ranking correlations are Spearman $\rho$ against the full 1,000-item ranking.}
\label{tab:robustprofile}
\centering
\begin{minipage}[t]{0.49\textwidth}
\centering
\scriptsize
\textbf{Human-confirmed sensitivity}\\[2pt]
\setlength{\tabcolsep}{2.3pt}
\renewcommand{\arraystretch}{0.96}
\begin{tabular}{@{}lrrrrr@{}}
\toprule
Subset & $N$ & A1 & A3 & $\rho$ & Ret. \\
\midrule
Full set & 1000 & .683 & .892 & -- & 8/8 \\
At least one & 950 & .691 & .897 & 1.000 & 8/8 \\
Both humans & 611 & .706 & .901 & .994 & 8/8 \\
\bottomrule
\end{tabular}
\end{minipage}\hfill
\begin{minipage}[t]{0.49\textwidth}
\centering
\scriptsize
\textbf{Higher-order profile}\\[2pt]
\setlength{\tabcolsep}{2.3pt}
\renewcommand{\arraystretch}{0.96}
\begin{tabular}{@{}lrrr@{}}
\toprule
Orientation & Ref. & Pred. & $\Delta$pp \\
\midrule
Openness to Change & 30.13 & 35.04 & +4.91 \\
Self-Enhancement & 19.95 & 19.81 & -0.14 \\
Conservation & 29.91 & 26.92 & -3.00 \\
Self-Transcendence & 20.00 & 18.23 & -1.78 \\
\bottomrule
\end{tabular}
\end{minipage}
\end{table}

\paragraph{Semantic mechanisms and source controls.}
The structured audit of 62 high-consensus texts explains the recurring transitions: universal scope contracts to interpersonal care; rule-following is replaced by its safety outcome; inherited practice becomes obedience; protective control is read as dominance; autonomous process eclipses achievement goals; and pleasure cues eclipse novelty or arousal. The mechanisms concern beneficiary scope, authority, agency, outcome, and reward rather than isolated lexical cues.

\begin{table}[H]
\caption{Structured audit of high-consensus cases. A case is included when at least 10 of 20 checkpoints produce the transition.}
\label{tab:caseaudit}
\centering
\scriptsize
\setlength{\tabcolsep}{3.0pt}
\renewcommand{\arraystretch}{0.95}
\begin{tabular}{@{}lrl@{}}
\toprule
Transition & Cases & Dominant mechanism \\
\midrule
Universalism$\rightarrow$Benevolence & 23 & Universal scope contracts to interpersonal care \\
Conformity$\rightarrow$Security & 11 & Norm or means is replaced by the safety outcome \\
Security$\rightarrow$Power & 9 & Protective control is interpreted as dominance \\
Stimulation$\rightarrow$Hedonism & 9 & Pleasure cues eclipse novelty or arousal \\
Tradition$\rightarrow$Conformity & 8 & Inherited practice is reduced to obedience \\
Achievement$\rightarrow$Self-Direction & 2 & Autonomous process eclipses the achievement goal \\
\bottomrule
\end{tabular}
\end{table}

Source and length controls do not account for the pattern. TAPE is descriptively easier, but its controlled estimate is positive and imprecise (OR $=1.76$, 95\% CI [0.88, 3.54], $p=.111$); standardized length is unrelated to correctness (OR $=1.05$, $p=.670$). Full coefficients are available in the linked repository.

\paragraph{Higher-order profile distortion.}
Strict higher-order accuracy is 0.820, yet 55.5\% of semantic top-1 disagreements cross a higher-order boundary. Fine-grained errors produce a net profile shift: the pooled predicted profile overstates Openness to Change by 4.91 percentage points and understates Conservation by 3.00 points and Self-Transcendence by 1.78 points. Systems that aggregate predicted values into user, organization, or corpus profiles can consequently inherit a directional bias even when many individual mistakes concern neighboring categories.

\section{Discussion}

\paragraph{Recognition and Ranked Recovery.}
The gap between Acc@1 (0.683) and Acc@3 (0.892), together with 87.5\% rescue for adjacent errors, shows that LLMs often locate the correct motivational neighborhood but rank neighboring values unstably. Evaluation should report ranked recovery alongside exact recognition. At the same time, Universalism, Conformity, and Tradition remain distinctly difficult, and the under-use of Universalism and Tradition shows that category-specific recall cannot be inferred from overall accuracy.

\paragraph{Shared Boundaries, Checkpoint-Specific Fingerprints.}
The eight robust transitions define a small shared error structure, but two mechanisms must be separated. Stimulation--Hedonism reflects bidirectional boundary ambiguity; Universalism$\rightarrow$Benevolence, Tradition$\rightarrow$Conformity, and Conformity$\rightarrow$Security are asymmetric category collapses. Checkpoints differ sharply in which collapse dominates, while family membership adds little explanatory similarity. Claims about ``what LLMs confuse'' require both cross-checkpoint replication and model-specific fingerprints.

\paragraph{Human Robustness and Practical Meaning.}
Independent support for 95.0\% of items and preservation of every robust transition on both confirmed subsets strengthen the measurement claim. The case audit further shows that errors concern scope, beneficiary, authority, means versus outcome, agency, and reward, suggesting target-aware prompts, contrastive examples, and secondary labels for genuine overlap. Because over half of semantic errors cross a higher-order boundary, local confusions can systematically bias aggregate value profiles.

\section{Limitations}

The evaluation uses one Russian names-only forced-choice prompt; definitions, demonstrations, multilingual prompts, and multilabel instructions may yield different boundary structures. Each annotation records one primary value, leaving secondary values, confidence, evidence spans, and the holder or target of the expressed value for future extensions. Exact construction-model agreement emphasizes clear examples, so natural prevalence and ambiguity may differ. Source composition is non-uniform, although source-controlled and human-confirmed analyses preserve the main results. The higher-order analysis assigns the boundary value Hedonism to Openness to Change under an explicit non-overlapping convention. Leave-one-family-out tests reduce, but cannot eliminate, dependence among related checkpoints. Downstream profile effects are estimated from observed error pathways and motivate direct deployment studies.

\section{Conclusion}

Primary-value recognition exhibits a reproducible directional structure. On 1,000 unique Russian texts with complete two-annotator validation, pooled Acc@1 is 0.683 and Acc@3 is 0.892; most adjacent disagreements retain the reference within the top three. Eight directed transitions recur across checkpoints and human-confirmed subsets, with severity determined more by the checkpoint than by the model family. The errors reflect semantic shifts in scope, authority, agency, outcome, and reward. More than half cross a higher-order boundary and collectively shift the inferred profile toward Openness to Change. Reliable value-recognition evaluation should combine exact and ranked metrics, controlled directed confusions, independent validation, model-specific fingerprints, and analysis of each systematic shift.

\section*{Data and Code Availability}
The evaluation code and paper materials are available in the
\href{https://github.com/ikanam-ai/VALAR}{public VALAR repository}.
The complete LLM-run data are archived in the
\href{https://osf.io/u56kq/overview?view_only=1c3bc242d37247de83e92113d7837be3}{OSF run archive}.

\end{document}